\documentclass[11pt,a4paper]{article}
\usepackage{emnlp2021}
\usepackage{times}
\usepackage{latexsym}
\usepackage{amsfonts,amssymb}

\usepackage{microtype}
\usepackage{placeins}
\usepackage{booktabs}
\usepackage{appendix}

\usepackage{svg}
\usepackage{amsmath}

\usepackage{bbm}
\usepackage{graphicx}  
\usepackage[utf8]{inputenc}
\usepackage{booktabs}
\usepackage{url}
\usepackage{multirow}
\usepackage{soul}
\usepackage{paralist}
\usepackage{enumitem}

\title{
A Token-level Reference-free Hallucination Detection Benchmark for Free-form Text Generation}

\author{Tianyu Liu$^1$ $^2$\thanks{\ \ Work was done when Tianyu (intern) and Yizhe was at Microsoft.}\ \ \ Yizhe Zhang$^3$  \ \ \ Chris Brockett$^4$ \ \ \  Yi Mao$^4$  \\ 
\textbf{Zhifang Sui$^1$ \ \ \ Weizhu Chen$^4$ \ \ \ Bill Dolan$^4$}\\
 $^1$ Peking University \ \ $^2$ Tencent Cloud Xiaowei  \ \ $^3$ Meta AI \ \ $^4$ Microsoft Corporation\\
 \texttt{\{tianyu0421,szf\}@pku.edu.cn} \ , \ 
 \texttt{yizhe.zhang@hotmail.com} \\ \texttt{\{chrisbkt,maoyi,wzchen,billdol\}@microsoft.com}  }

\date{}

\begin{document}

\maketitle

\begin{abstract}
Large pretrained generative models like GPT-3 often suffer from hallucinating non-existent or incorrect content, which undermines their potential merits in real applications.
Existing work usually attempts to detect these hallucinations based on a corresponding oracle reference at a sentence or document level.
However ground-truth references may not be readily available for many free-form text generation applications, and sentence- or document-level detection may fail to provide the fine-grained signals that would prevent fallacious content in real time. 
As a first step to addressing these issues, we propose a novel \emph{token-level, reference-free} hallucination detection task and an associated annotated dataset named \textsc{HaDes} (\textbf{HA}llucination \textbf{DE}tection data\textbf{S}et)
\footnote{Code and data are provided in \url{https://github.com/microsoft/HaDes}}. To create this dataset, we first perturb a large number of text segments extracted from English language Wikipedia, and then verify these with crowd-sourced annotations. To mitigate label imbalance during annotation, we utilize an iterative model-in-loop strategy. We conduct comprehensive data analyses and create multiple baseline models.
\end{abstract}

\section{Introduction}
 Automatic text generation using neural natural language generation (NLG) systems is increasingly fluent and thus seemingly plausible in many real-world applications. Large-scale pretrained models like GPT-3 \cite{brown2020language} are proven to be powerful in understanding and performing free form text generation tasks at human-quality level with a few in-context examples, which dramatically reduces the manual labor needed in many text-based applications and services. 
Despite their great success, however,
neural NLG systems using very large pre-trained models struggle to generate factually accurate and trustworthy text \cite{devlin-etal-2019-bert,radford2019language}, and exhibit a propensity to hallucinate non-existent or incorrect content that is unacceptable in most user-oriented applications. This poses a major challenge for deploying production NLG systems with realtime generation, where post-examination is impossible. 

Existing work has sought to detect hallucination and quantitatively measure generation consistency against a provided reference. Such \emph{reference-based} hallucination detection has been proposed for abstractive summarization \cite{maynez-etal-2020-faithfulness}, machine translation \cite{wang-sennrich-2020-exposure}, data-to-text generation \cite{rebuffel2021controlling}, and image caption generation \cite{rohrbach-etal-2018-object}. 
For many free-form text generation tasks, however,
references are not readily available. 
For example, in a production NLG system such as a social chatbot using real-time response generation or a document auto-completion system, the generation model often cannot pair its outputs with sufficient reference information, rendering reference-based methods less applicable: $i)$ It may be difficult to even know where to obtain the reference, as obtaining it may be as hard as generating consistent information in the first place; 
$ii)$ Generation may be at a real-time online setting that demands leveraging only existing context to create new content.

One common setup for qualitatively measuring the level of hallucination is performed at \textit{sentence- or document-level} \cite{dhingra-etal-2019-handling,scialom2019answers}. Related tasks such as fake news detection \cite{zellers2019neuralfakenews} or fact checking \cite{thorne-vlachos-2018-automated}  also adopt this strategy. 
However, sentence- or document-level detection may not always provide high-resolution signals sufficient to pinpoint the hallucinated text, or can only judge whether a generated sentence or a document as a whole is a hallucinated artifact. Consequently, these high-level strategies may be insufficient to avoid hallucinations. As an alternative, at decoding time of an NLG system, we suggest that if the locus of hallucination can be identified at the token level, it may be possible to guide beam search or suppress the probability of certain tokens at real-time.

To this end, 
we propose a \emph{reference-free,  token-level hallucination detection} task and introduce an annotated training and benchmark testing dataset that we call \textsc{HaDes} (\textbf{HA}llucination \textbf{DE}tection data\textbf{S}et). 
The \textit{reference-free} property of this task yields greater flexibility in a broad range of generation applications. We expect the \emph{token-level} property of this task to foster the development of models that can detect fine-grained signals of potential hallucination.  
In conjunction with consulting context to identify self-contradictory statements and access to commonsense and world knowledge, such fine-grained signals, when detected, should further mitigate real-time hallucination.

Our contributions include:
\textbf{1)} We propose a \textit{reference-free},  \textit{token-level} hallucination detection task for free-form text generation.
\textbf{2)} We support this task with a dataset that we call \textsc{HaDes}, with $\sim$11k instances extracted from English Wikipedia using an iterative data collection strategy to address data imbalance issues. We also present comprehensive analyses on the statistical features to shed light on what is commonly recognized as hallucination in crowd-sourced judgments and its salient characteristics in free-form text generation.
\textbf{3)} We create multiple baselines, including feature based models and pretrained models as a first step towards addressing the proposed task.

\begin{figure}
    \centering
\includegraphics[width=0.8\linewidth]{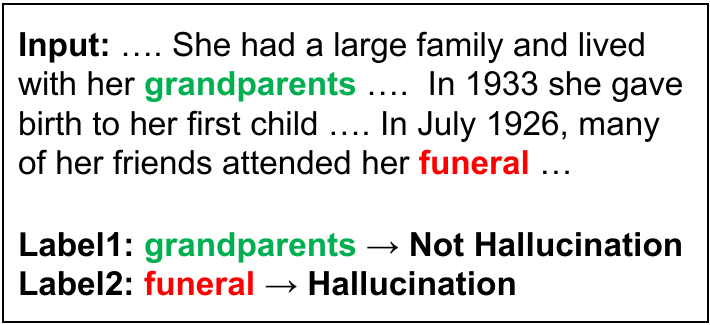}
    \caption{Overview for reference-free token-level hallucination detection task.}
    \label{fig:task_overview}
\end{figure}

\section{Task Overview}
\label{sec:task_overview}

We formulate our hallucination detection task as a binary classification task. As shown in Fig~\ref{fig:task_overview}, our goal is to assign either a ``hallucination''\footnote{“Hallucination” in our paper refers to certain types of mistakes (Fig \ref{fig:hallu_type}) made by the NLG models. The notions of “consistency” and “not hallucination” are only for annotation purposes (Sec \ref{sec:data_annotation}).}(abbreviated as ``$\mathcal{H}$'')
or a ``not hallucination'' (abbreviated as ``$\mathcal{N}$'') label to the highlighted spans. 

To simulate real-world NLG applications, we propose two sub-tasks with ``\textit{offline}'' and ``\textit{online}'' settings. In the \textit{offline} setting, it is assumed that generation is complete, so the the model is able perceive the bidirectional context. This could be used in the post-generation examination of NLG systems. 
For \textit{online} detection, the model can only access the unidirectional preceding context, which simulates on-the-fly generation. Online detection is important in practice as it enables NLG systems to proactively forestall potential hallucinations. 

\begin{figure}[t!]
\begin{center}
\includegraphics[width=1.0\linewidth]{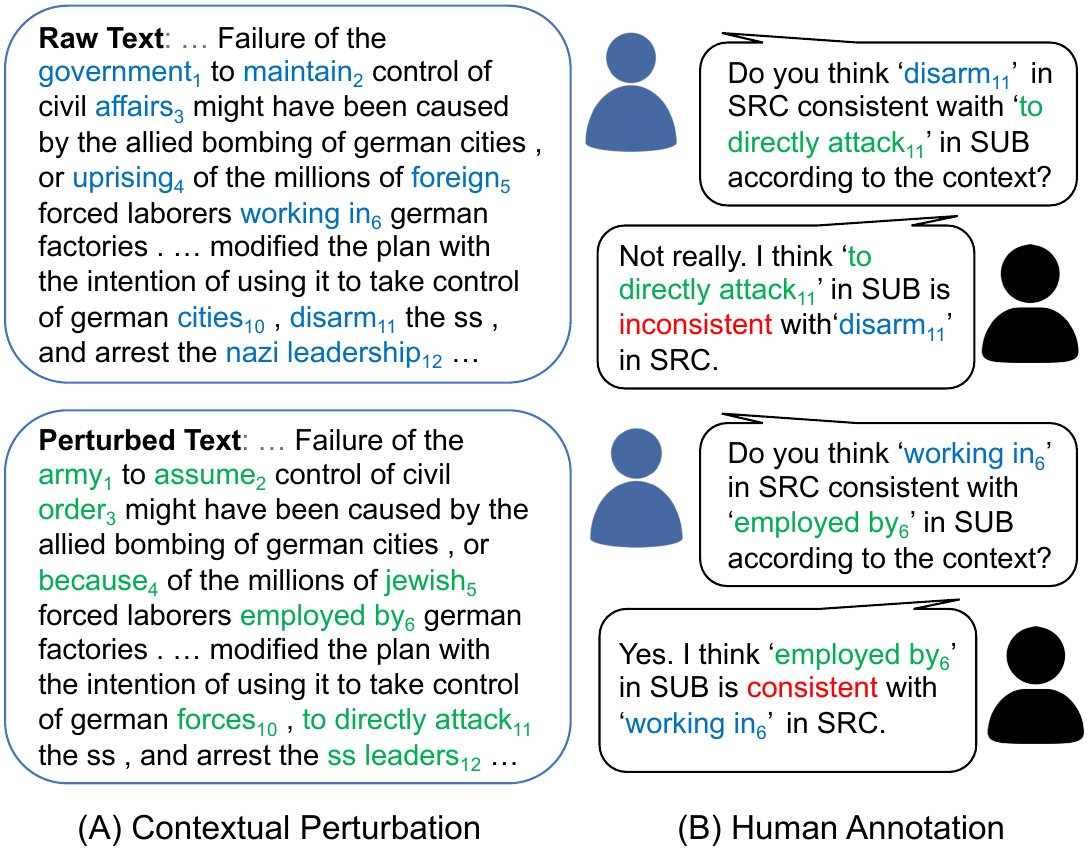}
\end{center}
\caption{The data collection process of \textsc{HaDes}.}
\label{fig:task-overview}
\end{figure}

\section{Dataset Creation}

To collect the \textsc{HaDes} dataset, we first perturb ``raw text'' web data into ``perturbed text'' (Fig \ref{fig:task-overview}A) (Sec \ref{sec:contextual_perturbation}). We then ask human annotators to assess whether the perturbed text spans are hallucinations given the original text 
(Fig \ref{fig:task-overview}B) (Sec \ref{sec:data_annotation}). 

\subsection{Raw Data Collection}
Our raw data are sampled from English \textsc{Wiki-40B} \cite{guo-etal-2020-wiki} dataset. \textsc{Wiki-40B-En} is a cleaned collection of English Wikipedia articles. We randomly sample from the first paragraphs of these articles and filter out short text of fewer than 5 sentences. We use Wikipedia as our text source since it is stylistically formal and of high quality, and covers diverse topics and domains. 

\subsection{Contextual Perturbation}
\label{sec:contextual_perturbation}

To acquire machine generated text in the free-form, we perturb the raw text \footnote{In a pilot study, we tried to annotate a token-level dataset based on GPT-3 generated text. However, we found that annotators had trouble achieving consensus if we don’t provide the “original text”. The size of the resulting data would be small. We thus reduce the ambiguity and subjectivity in the annotation process by asking if the pinpointed position in perturbed text is consistent/hallucinated compared with the original reference text. }
using BERT. In applying this contextual perturbation we maintained two principles: $i)$ the fluency and syntactic correctness of the perturbed text should be preserved; $ii)$ the perturbed text should be lexically diverse. 

We leave the first two sentences in the raw text unchanged to serve as the preceding context, so as to avoid the ``early token curse'' \cite{press2020shortformer} where tokens are evaluated at the beginning with limited context. The text perturbation process is split into three pipelined operations, namely \texttt{MASK}, \texttt{REPLACE} and \texttt{RANK}.
\begin{itemize}
    \item \textbf{i)} In the \texttt{MASK} operation, we mask the tokenized words to be replaced with the special token ``[MASK]'' in the BERT vocabulary. Starting from the third sentence, we randomly mask word spans by a pre-defined mask ratio $\rho$. By default we only mask one word in each perturbation, except for named entities identified by \emph{Spacy}. We view the entity boundaries as minimal masking units to avoid collocation errors (\textit{e.g.} ``San Diego'' should be masked as a whole). To reduce trivial instances, we do not mask stop words or punctuation identified by NLTK \cite{bird2006nltk}.
    
    \item \textbf{ii)} In the \texttt{REPLACE} operation, we leverage a pretrained BERT-base model to predict the masked span. The mask-then-predict training framework of the BERT model contextualizes the replacement with both preceding and subsequent text. For better fluency, we replace the masked tokens from left to right, \textit{e.g.} a 3-token \texttt{REPLACE} operation will be ``[MASK] [MASK] [MASK]'' $\to$ ``[A] [MASK] [MASK]'' $\to$ ``[A] [B] [MASK]'' $\to$ ``[A] [B] [C]''\footnote{It is possible to substitute the original tokens with more or fewer of tokens. However enumerating all possible token lengths is difficult, and empirically we see marginal gain in diversity in the resulting perturbed text. In our experiments we use same number of tokens for replacement.}. When performing the replacement, we remove the \textit{original} token from the predicted distribution over the vocabulary at each position of the text span, to avoid duplicated text after perturbation.
    We compared several decoding strategies in token substitution, including greedy, top-k (k=5/10/50) and top-p (p=0.95/0.9/0.8) \cite{holtzman2019curious} sampling methods. For comparison we sample 30 perturbed text for each sampling method and 
    count the number of incoherent perturbations. We choose top-k (k=10) sampling as its good trade-off between diversity (via number of distinct tokens) and coherence (via number of incoherent perturbations).
    
    \item \textbf{iii)} For each perturbed text, we substitute multiple word spans. Although being locally coherent, the perturbed text may still exhibit some global incoherence and syntactic issues, especially for longer text.
    We thus post-process the perturbed text with a \texttt{RANK} operation as an additional screening step. For each raw text, we generate 20 perturbed candidates and rank them according to language model perplexity using a GPT-2 (117M) model. We only keep the the candidate with lowest perplexity to ensure the fluency and syntactic correctness. 
\end{itemize}



\subsection{Data Annotation}
\label{sec:data_annotation}
We ended up with $\sim$1M perturbed text segments in the pool after contextual perturbation, not all of which contain hallucination, as the BERT model can generate factual information given that it is pretrained on a rich open web corpus. Thus, we sought to further annotate the automatically perturbed texts via crowd-sourcing. Human annotation is prohibitively expensive at this scale, so instead of annotating all 1M perturbed texts, we annotated a subset that is  \textit{less trivial} and would lead to a more \textit{balanced} distribution, using an iterative model-in-the-loop annotation approach that is conceptually related to active learning \cite{cohn1996active,jia-liang-2017-adversarial,zellers-etal-2018-swag,nie-etal-2020-adversarial}.


\paragraph{Human annotation settings}
 To perform the annotations, 
 we hired judges on an internal (the name is redacted for double-blind review) crowd-sourcing platform comparable to AMT. 
 The judges were limited to the North American English speakers with good records (recognized as experts in the platform, rejection rate $\leq$ 1\%)  and were screened via a simple 10-question qualification test (answering 8 out of 10 questions correctly).
 They were paid 0.15\$ per HIT, which is more than prevailing local minimum wage.
 Protocols were implemented to block spammers in real time \footnote{If a worker keeps choosing the same label for all HITs, or the average time spent per HIT is less than 10 seconds, or more than 30\% of their judgments conflict with others', we would manually check their annotations and block the spammers.}.
For each annotation, both original text and perturbed text were shown to the judges, with perturbed text span highlighted. The annotators were asked to determine whether the perturbed text spans are $\mathcal{H}$ (hallucination) or $\mathcal{N}$ (not hallucination) with the original text in terms of factualness and semantic coherence given the context.
Each pair was judged by 4 annotators, and up to 6 if consensus was not reached. We retained only those annotations for which consensus was reached.
Out of 12,719 annotated instances, 86.12\% instances reach consensus and are included in \textsc{HaDes} dataset; 78.47\% instances reach $\geq$ 80\% agreement among annotators, e.g. 4/5 or 5/6 vote for ``hallucination'' label; 71.24\% instances reach 100\% agreement in the annotation.
For inter-annotator agreement (IAA), the Krippendorf's alpha between the annotators is 0.87.



\paragraph{Iterative Model-in-the-loop annotation}
Annotating all perturbed text segments is expensive and time-consuming. Thus, we resort to annotating a subset. We applied two principles for selecting the data to be annotated: $i)$ the data should be \textit{balanced}. We found that with randomly sampled instances, the annotated label distribution is heavily skewed toward the ``hallucination'' class. Presumably most contextualized perturbations result in factual inconsistency to certain extent. However, we aim to have the number of instances in both classes on par with each other, so that the ROC (receiver operating characteristic) curve of tested models can be better characterized.  
$ii)$ the data for annotation should be \textit{less trivial} \footnote{Many perturbations are \textit{trivial} to predict, \textit{e.g.} replacements that change a specific date to a non-date-related phrase must be a hallucination.}. The obvious instances contribute little to model training and method benchmarking, but cost as much annotation effort as other instances. 

The challenge is that we cannot know \textit{a priori} the annotation labels and ease of labeling, hence selecting \textit{less trivial} instances and forming a \textit{balanced} label distribution for annotation is not straightforward.
To address this challenge, we adopt an iterative \textit{Model-in-the-loop} annotation strategy. Specifically, we split the annotations into several rounds. For each round \footnote{Except the first round, where we use random sampling.}, we first retrain a hallucination detection model (initiated with BERT) based on the annotated instances in the previous rounds. This model is used for selecting the next batch of data to be annotated from the remaining unlabeled data. 

To filter out trivial instances and focus on the more \textit{useful} cases, we use a heuristic rule for the automatic screening by abandoning instances where the detection model assigns low or high probability to ``hallucination'' class (the threshold varies in different rounds to yield reasonable number of candidates). To eliminate cases where the perturbed text paraphrases the original text, we also measured the cosine similarity between the replaced text (through ``[CLS]'' representation) and corresponding original content using a RoBERTa model (without fine-tuning), and then filtered out cases with a similarity score greater than 0.9. We also remove a large portion of obvious hallucination instances where the target text span is recognized as a \textsc{Date} or \textsc{Name}, and replaced by a different \textsc{Date}\footnote{We only remove cases where the replaced date is \textit{definitely} different (e.g., from ``Monday'' to ``Tuesday''). We do not remove ambiguous cases such as from ``today'' to ``Tuesday''.} or \textsc{Name}.

In the initial rounds of annotation, we observed extreme label imbalance (around 90\% are $\mathcal{H}$ class) between $\mathcal{H}$ (hallucination) and $\mathcal{N}$ (not hallucination) cases. 
To rebalance the label distribution so that each class received a decent amount of annotation, we performed additional subsamping based on the label predicted by the aforementioned detection model. 
We assume the human annotation for $\mathcal{H}$ and $\mathcal{N}$ cases is the oracle, indicating actual $\mathcal{H}$/$\mathcal{N}$. Since the actual ``hallucinated'' is dominant, we seek to subsample from instances that are predicted as $\mathcal{H}$ by the detection model to make the distribution of actual $\mathcal{H}$/$\mathcal{N}$ even. To do this, we estimate the true positive rate (TPR, $\alpha$), true negative rate (TNR, $\beta$) and true precision ($\gamma$) of the detection model based on the annotation from last round. The hope is that after subsampling, the actual $\mathcal{H}$ (TP + FN) is roughly equal to actual $\mathcal{N}$ (FP + TN). 
The estimated subsampling ratio $R$ for the predicted $\mathcal{H}$ (TP + FP) is given by\footnote{Details are provided in the appendix.}:
\begin{align}
    R = \frac{-2\alpha\beta\gamma + \alpha\beta + \beta\gamma + \alpha\gamma-\gamma}{(2\gamma-1)\alpha(1-\beta)}
\end{align}

\begin{figure*}[t!]
\begin{center}
\includegraphics[width=1.0\linewidth]{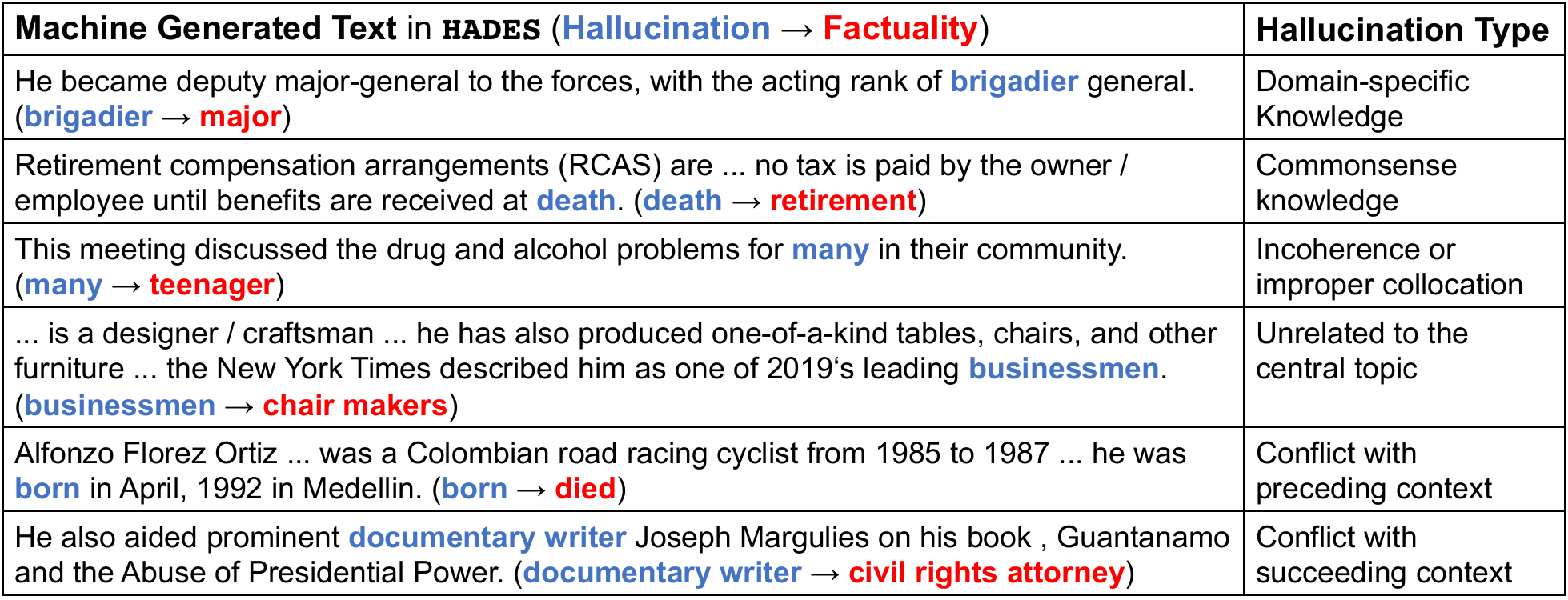}
\end{center}
\caption{Overview for different types of hallucination in the proposed \textsc{HaDes} dataset.}
\label{fig:hallu_type}
\end{figure*}

\subsection{Data Analysis}
\label{sec:data_analysis}
Below we provide data statistics and characterize the composition and properties of \textsc{HaDeS}.

\paragraph{Data statistics} In total, after accumulating annotations for several rounds, we obtain 12,719 instances with 71,226 HITS from judges. We conduct 14 rounds of annotation, increasing the annotation scale with each round (ranging from $\sim$200 instances/round to $\sim$4000 instances/round). Out of 12,719 annotated instances, 10,954 instances reached consensus among judges and are included in the \textsc{HaDes} dataset. We split the dataset into train, validation and test sets with sizes of 8754, 1000, 1200 respectively. In the final dataset, ``hallucination'' cases slightly outnumber ``not hallucination'' cases, with a ratio of 54.5\%/45.5\%. We summarize some typical hallucination types seen in the \textsc{HaDes} dataset in Fig \ref{fig:hallu_type}. 

\begin{figure*}[]
\begin{center}
\includegraphics[width=0.95\linewidth]{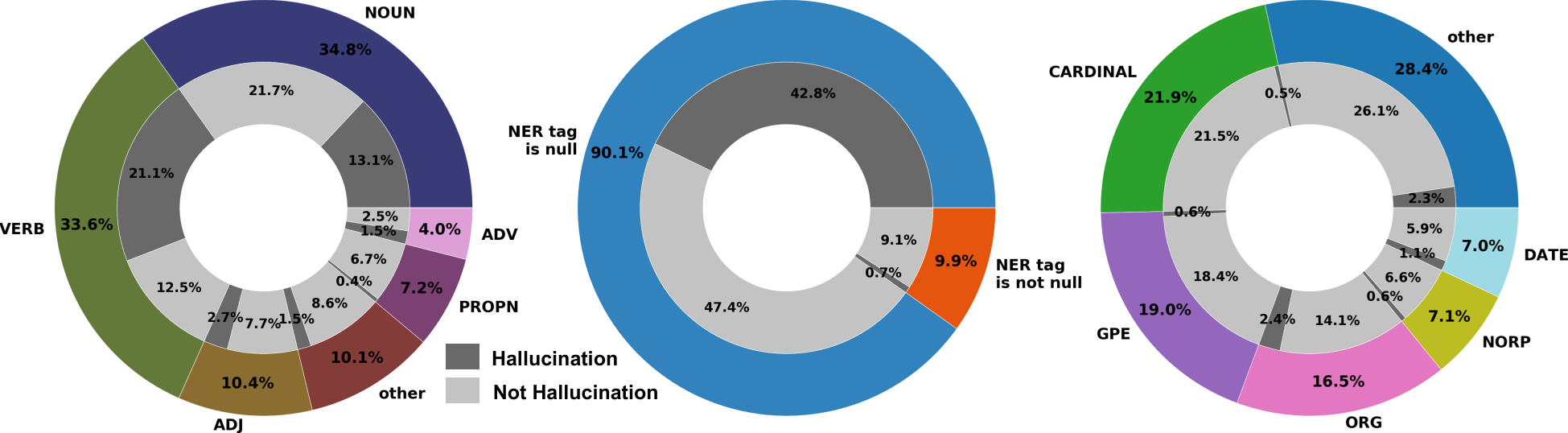}
\end{center}
\caption{Distributions of POS (left), NER (middle) and a breakdown of non-null NER tags (right) in \textsc{HaDes}.
}
\label{tab:stats_pos_ner}
\end{figure*}

\paragraph{Parsing features}
In Fig \ref{tab:stats_pos_ner} we show the ratio of ``hallucination''($\mathcal{H}$)/ ``not hallucination'' ($\mathcal{N}$) cases for different Part-of-Speech (POS) and Name Entity Recognition (NER) tags, identified by \emph{Spacy}.
From a POS perspective, around two-thirds of verbs and verbal phrases in the dataset are identified as ``not hallucination'', while in other types of words/phrases, ``hallucination'' cases are in the majority, e.g., most adverbs (ADV), adjectives (ADJ) and acronyms of proper nouns (PROPN) are labeled as ``hallucination''. Presumably many verbs or verbal phrases are lower in word concreteness \cite{nelson1992word} than other word types (\textit{e.g.} ``make'' and ``create'' can be used interchangeably in many circumstances), and thus,  as we observe in our dataset, are less prone to be perturbed into hallucinations.
For NER tags, about 90\% of word spans are not recognized as name entities. However, of the 10\% of remaining instances, over 90\% are ``hallucination'' cases. 


\begin{table}[ht!]
\centering
\small
\begin{tabular}{l|c|c|c|c}
\hline
Label & Word Prob$^*$ & Entropy & TF-IDF & PPMI \\\hline
$\mathcal{H}$ &  $\mathbf{5.85}_{25.6}$ & $\mathbf{2.58}_{1.49}$ & $\mathbf{.021}_{.019}$ & $.198_{.134}$ \\
$\mathcal{N}$ & $1.30_{7.67}$ & $1.78_{1.07}$ & $.019_{.014}$ & $\mathbf{.216}_{.129}$\\\hline
\end{tabular}
(A) $\mathbf{Mean}_{\mathbf{std}}$ statistics for \textbf{H}allucination ($\mathcal{H}$) and \textbf{not H}allucination ($\mathcal{N}$) labels (* indicates $\times1e^{-8}$).
\includegraphics[width=0.95\linewidth]{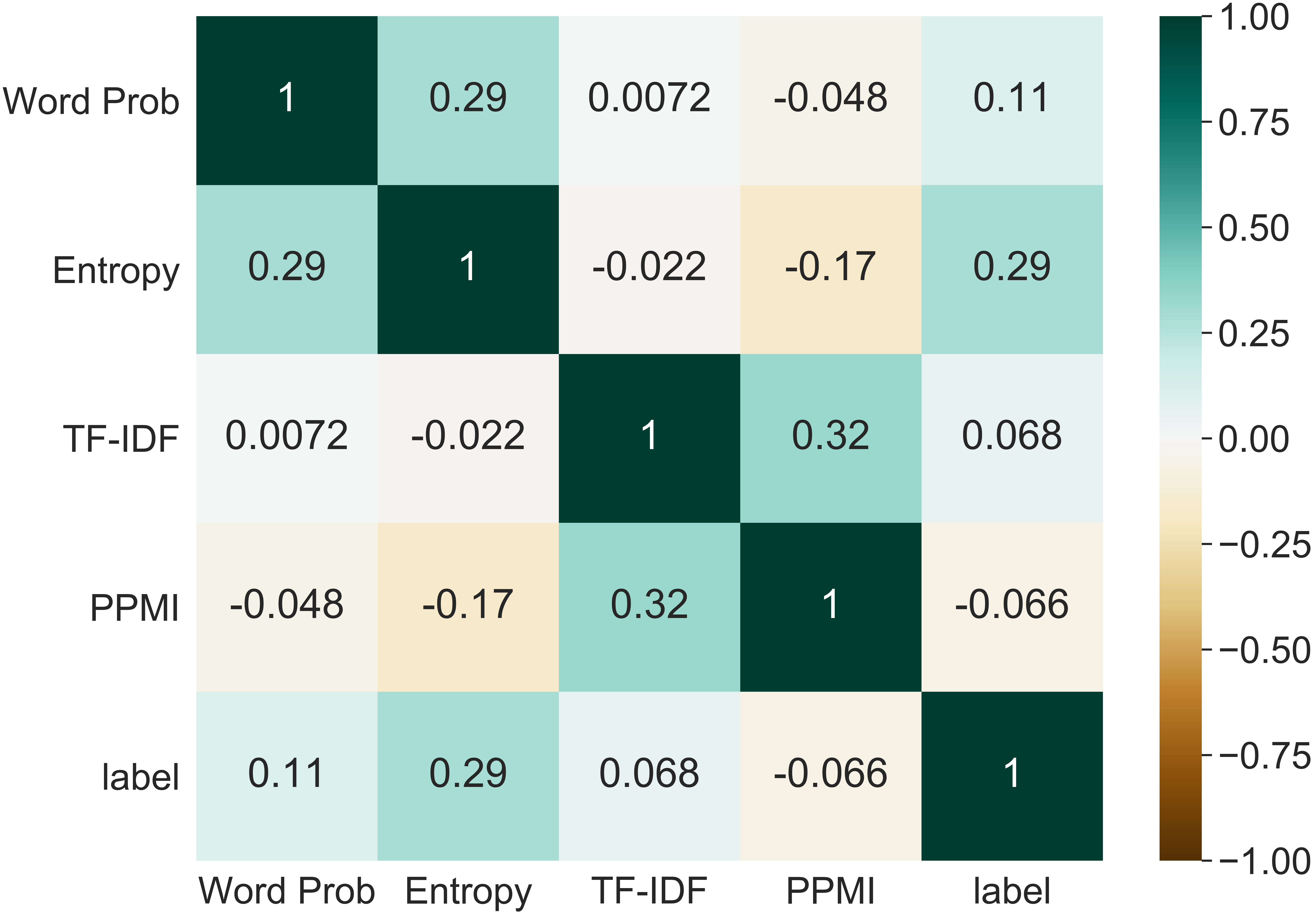}
(B) Feature correlation heatmap between hallucination label and word probability, entropy, TF-TDF and PPMI.    
    \caption{Analysis for statistical and model-based features of  \textsc{HaDes}.}
    \label{tab:stats_feature}
\end{table}

\paragraph{Statistical and model-based features}
To analyze the characteristics of hallucinations in \textsc{HaDes}, we compute the correlation between a selected group of statistical/model-based features and hallucination labels. As shown in Table \ref{tab:stats_feature}\footnote{More statistical feature analysis is in the appendix.}
, we obtain the average word probability and average word entropy of a given text span with a BERT base model (without fine-tuning), as well as term frequency–inverse document frequency (TF-IDF), positive pointwise mutual information (PPMI) features of the given word span. By comparing the features of the two labels ($\mathcal{H}$/$\mathcal{N}$) (Table \ref{tab:stats_feature}A), we observe that in our dataset, hallucinations typically associate with higher entropy. 
A counter-intuitive observation is that the hallucinations tend to have higher average probability than factually consistent content. We presume the underlying reason might be that the word distribution generated by machine may diverge from the word distribution of real human-written text \cite{holtzman2019curious,see2019massively} owing to self-reinforcing the current generation based on previous generation. Consequently, many overconfident generation outputs are likely to fall into hallucination. We observe no strong correlation between hallucination labels and TF-IDF or PPMI as demonstrated in Table \ref{tab:stats_feature}B.

\section{Baseline Models}
\label{sec:baseline_models}
As an initial step towards tackling the proposed hallucination detection task and benchmarking methods, we create several baseline detection models\footnote{The proposed token-level, reference-free hallucination detection hasn’t been covered in the existing literature. Thus this thread is first-of-its-kind. We are unable to find a feasible baseline that perfectly fits in our setting, therefore we propose multiple feature-based/pretrained baselines.}.

\paragraph{Feature-based models} As elaborated in Sec \ref{sec:data_analysis}, the statistical/model-based features like average word probability, average entropy, TF-IDF, PPMI, as well as parsing features like POS and NER tags can be vague indicators of hallucinations. The former two are context-aware and the latter four are not. We incorporate them as features to build classifiers including logistic regression (\textbf{LR}) and support vector machine (\textbf{SVM})  using \emph{scikit-learn} \cite{pedregosa2011scikit}. 
The maximum number of iteration is set as 100, with an early-stop strategy which stops training if the loss does not drop within 5 iterations. 

\paragraph{Transformer-based models} We also build baseline detection models based on pretrained transformer models including BERT, GPT-2, XLNet \cite{yang2019xlnet} and RoBERTa \cite{liu2019roberta}. These transformer-based models represent the state-of-the-art, and can potentially better leverage context or embedded world knowledge to detect self-contradictory or anti-commonsense content. 

Specifically, for an input text segment, we finetune a pretrained model $\mathcal{M}$ to predict binary hallucination labels  $\mathbf{y}$ for each given text span. During inference time, from the last layer hidden states $\mathbf{H} \in \mathbb{R}^{l \times h}$ ($h,l$ are hidden size and sequence length, respectively) of $\mathcal{M}$, suppose the target text span starts at position $s$ and ends at position $t$, we first obtain the representation $\mathbf{w} \in \mathbb{R}^{h}$ for the target span with max pooling (\textit{i.e.}, $\mathbf{w}= \mathrm{max\_pool}(\mathbf{H}_{s:t})$). 
We then map $\mathbf{w}$ to a binary hallucination label $y \in \{0,1\}$ with a MLP network using $\mathrm{tanh}$ as activation.
During training time, we fine-tune the model using cross entropy objective between the predicted labels and the actual labels.




\begin{table*}[t]
    \centering
\begin{tabular}{lcccccccccc}
    \hline
    \multirow{2}{*}{Model}  &  \multirow{2}{*}{Acc} &  \multirow{2}{*}{G-Mean ($\uparrow$)} &  \multirow{2}{*}{BSS ($\downarrow$)} &
    \multirow{2}{*}{AUC} & \multicolumn{3}{c}{Not Hallucination} & \multicolumn{3}{c}{Hallucination} \\ 
    & & & & & P & R & F1 & P & R & F1 \\ \hline
    LR & 62.25 & 60.77 & - & - & 62.35 & 72.08 & 66.86 & 62.10 & 51.24 & 60.33\\
    SVM & 63.67 & 61.50 & - & - & 62.89 & 76.18 & 68.90 & 65.05 & 49.65 & 56.31 \\
    BERT  & 71.92 & \textbf{71.95} & 19.06 & 78.63 & \textbf{74.46} & 71.29 & 72.84 & 69.31 & \textbf{72.61} & 70.92\\
    RoBERTa  & \textbf{72.83} & 70.94 & \textbf{18.78} & 78.72 & 74.06 & 74.76 & 74.41 & 71.43 & 70.67 & \textbf{71.05}\\ 
    XLNet  & 72.33 & 71.39 & 18.79 & \textbf{78.93} & 71.15 & \textbf{80.13} & \textbf{75.37} & \textbf{74.07} & 63.60 & 68.44\\ 
    \hline

    \end{tabular}
    \caption{Benchmark (numbers in percentages (\%)) for the offline setting on \textsc{HaDes}, where detecting models have access to the bidirectional context. $\downarrow$/$\uparrow$ indicates lower/higher is better. Significant tests are in the appendix.}
    \label{tab:results_offline}
\end{table*}

\begin{table*}[t]
    \centering
    \begin{tabular}{lcccccccccc}
    \hline
    \multirow{2}{*}{Model}  &  \multirow{2}{*}{Acc} &  \multirow{2}{*}{G-Mean ($\uparrow$)} &  \multirow{2}{*}{BSS ($\downarrow$)} &
    \multirow{2}{*}{AUC} & \multicolumn{3}{c}{Not Hallucination} & \multicolumn{3}{c}{Hallucination} \\  
    & & & & & P & R & F1 & P & R & F1 \\ \hline
    GPT-2 & \textbf{71.58} & \textbf{70.98} & 19.13 & 77.71 & \textbf{71.32} & 77.29 & \textbf{74.19} & \textbf{71.93} & \textbf{65.19} & \textbf{68.40} \\
    BERT  & 71.00 & 70.43 & \textbf{18.66} & \textbf{78.83} & 70.91 & 76.50 & 73.60 & 71.12 & 64.84 & 67.84 \\
    RoBERTa & 70.67 & 70.14 & 19.77 & 77.07 & 70.74 & 75.87 & 73.22 & 70.58 & 64.84 & 67.59\\
    XLNet & 70.08 & 69.17 & 19.76 & 76.59 & 69.39 & \textbf{77.60} & 73.27 & 71.08 & 61.66 & 66.04  \\\hline

    \end{tabular}
    \caption{Benchmark (numbers in percentages (\%)) for the online setting on \textsc{HaDes}, where detection models only have the access to left context. $\downarrow$/$\uparrow$ indicates lower/higher is better. Significant tests are in the appendix.}
    \label{tab:results_online}
\end{table*}

\begin{figure*}[]
\begin{center}
\includegraphics[width=1.0\linewidth]{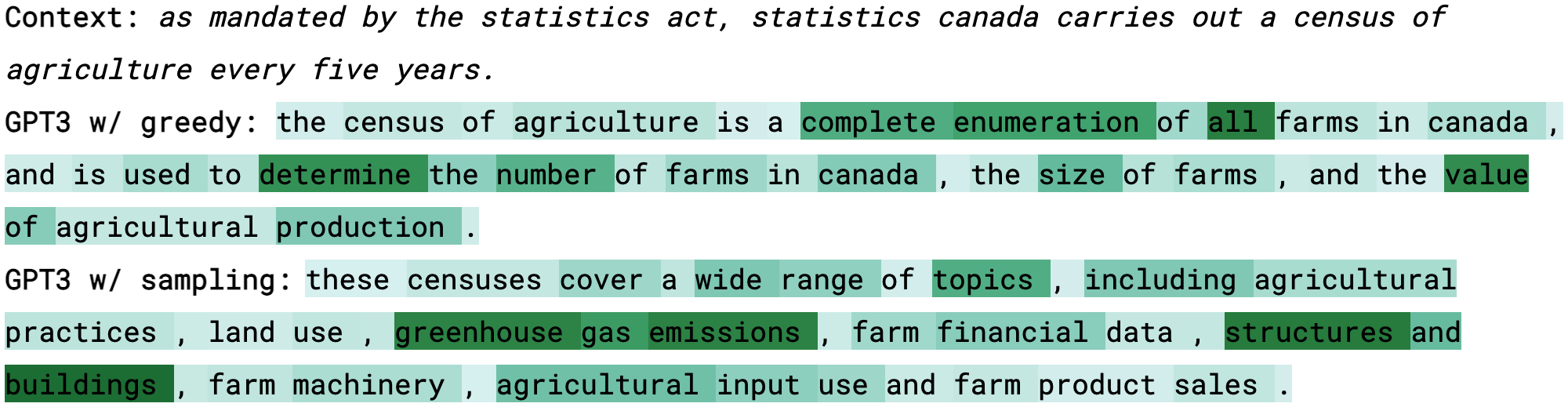}
\end{center}
\caption{The visualization of predicted hallucination scores for a sample of GPT-3 generated text, provided by BERT (large, offline) detector. Darker green signifies higher risk to be hallucinations.
}
\label{fig:model_viz}
\end{figure*}

\begin{figure}[t!]
\begin{center}
\includegraphics[width=1.0\linewidth]{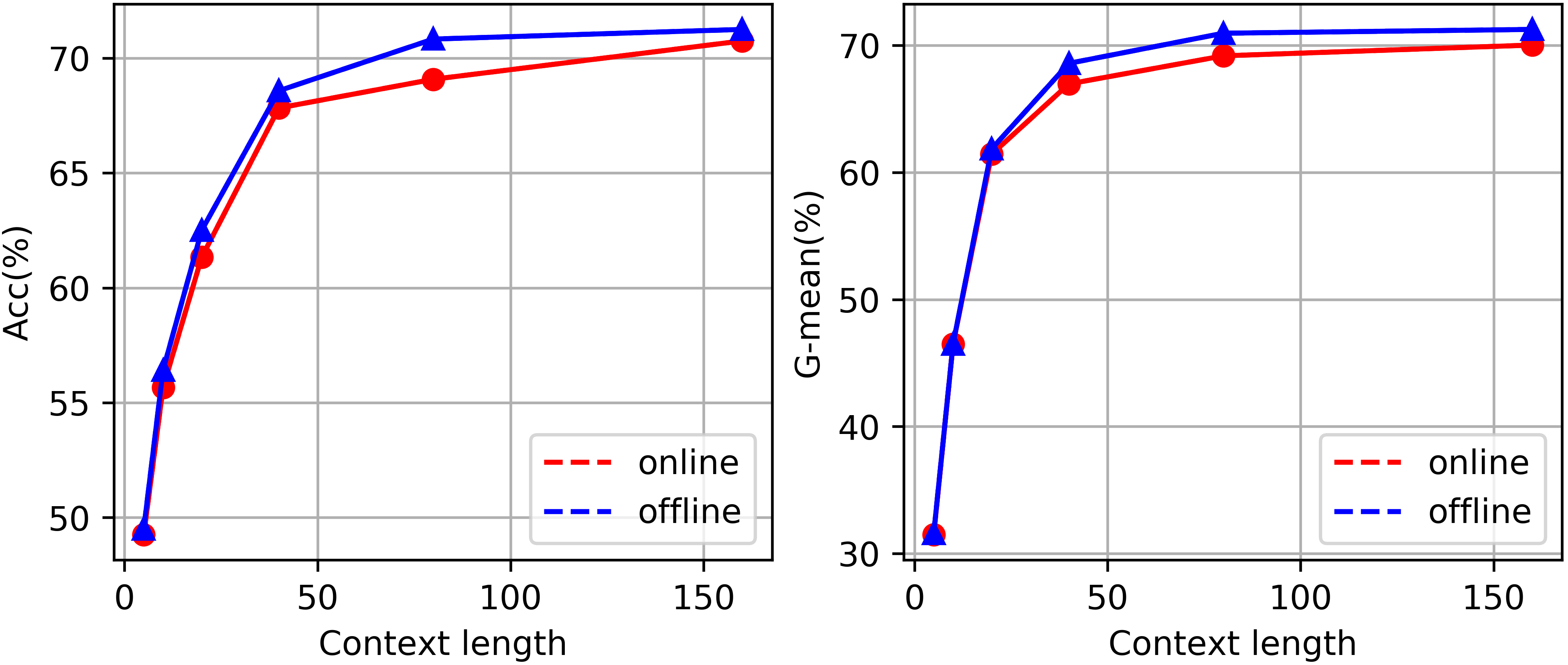}
\end{center}
\caption{The performance of BERT-large based detecting model with different context lengths.}
\label{fig:context_len}
\end{figure}

\section{Experimental Setup}

\paragraph{Baseline configurations}

For the transformer-based baselines, we experiment with a variety of pretrained models via Hugging Face Transformers \cite{wolf-etal-2020-transformers}, including \textbf{BERT}-large (335M), \textbf{GPT2}-medium (345M), \textbf{XLNet}-large (340M), \textbf{RoBERTa}-large (355M). We use Adam optimizer \cite{kingma2014adam} with different learning rates, i.e. 5e-3 for GPT2 and BERT and 1e-3 for other models.

We explored multiple model architectures and setups to determine the optimal configuration using BERT-large model. These include $i)$ span representation with mean/max pooling ; $ii)$ number of layers of the MLP network; $iii)$ hidden dimension of the MLP ; $iv)$ whether or not to freeze the parameters of $\mathcal{M}$ up to the last layer, and choose the best configuration according to model performance on the validation set. The best configuration uses max-pooling, employs 2 layers of MLP with hidden dimension of $h/2$, and freezes the model parameters up to the last layer of $\mathcal{M}$ and just fine-tunes the binary MLP classifier.
We apply the same network configuration to all other pretrained models as empirically we see marginal performance gain after enumerating different configurations for individual pretrained models other than BERT.

As discussed in Sec.\ref{sec:task_overview}, \textsc{HaDes} can serve as benchmark for hallucination detection in both offline (model can see bidirectional context) and online (only preceding context can be leveraged) settings.
Note that we apply the feature-based baselines only in the offline setting (Table \ref{tab:results_offline}), because a good estimation of those features requires bidirectional context. The transformer with causal attention (GPT-2) can only fit in the online setting.

\paragraph{Evaluation metrics}
We evaluate the baselines on HaDes with standard classification metrics including accuracy, precision, recall, F1 and AUC (Area Under Curve) with respect to ROC.
We also utilize the G-Mean metric which measures geographic mean of sensitivity and specificity \cite{espindola2005extending} and they were reported useful especially for the imbalanced label distribution scenarios. 
We also employ the Brier Skill Score (BSS) metric \cite{center2005brier}, which calculates the mean squared error between the reference distribution and the hypothesis probabilities.

\section{Results}
\paragraph{Baseline performance}
Table \ref{tab:results_online} and Table \ref{tab:results_offline} show the performance of the baseline models 
\footnote{To identify the clear winner among baseline models, we report the significant tests for the baseline models in Table \ref{tab:results_online} and Table \ref{tab:results_offline} as follows:
For the offline setting (Table \ref{tab:results_offline}), there is no obvious winner among pretrained models, e.g. RoBERTa wins in ACC; XLNet wins in F1 for not hallucination cases; BERT wins in G-mean.
For the online setting (Table \ref{tab:results_online} ), we ran significant tests for the mean performance (over 5 runs) between GPT-2 and BERT; GPT-2 and XLNet; GPT-2 and RoBERTa, the differences in terms of ACC; G-mean; F1 scores for both hallucination and not hallucination labels are significant (alpha=0.01) after Bonferroni correction.} 
in both online and offline settings respectively. In both settings, the predictions for ``not hallucination'' cases have higher F1 scores than ``hallucination'' cases. All models perform better in the offline setting compared with the online setting, indicating that the succeeding context of the target words helps identify hallucinations. The transformer-based baselines are generally on par with each other. 
Under the offline setting, the pretrained models outperform feature-based models by a large margin; this indicates that the powerful contextualized feature extractor is important for successfully identifying  hallucinations at fine granularity.
Under the online setting, we observe that, for most of the metrics, GPT-2 yields the best performance of all baselines. Presumably, the causal language model pretraining method makes GPT-2 perform better in the auto-aggressive (online) detection setting.
\paragraph{Context matters in \textsc{HaDes}} 
To investigate extent to which contextual information helps the hallucination detection in \textsc{HaDes}, we run BERT-large detection model with different context lengths and characterize its performance in both online and offline settings in Fig \ref{fig:context_len}. 
Starting from the target words, we set a fixed size (5/10/20/40/80/160) context window and truncate all text beyond this window. As we enlarge the context window, model performance grows rapidly when context length is smaller than 80, and then gradually converges. This observation highlights the importance of context in hallucination detection. Interestingly,  we observe that the model obtains higher performance in the offline mode than in the online setting. The performance gap between the two settings maximizes when context length is around 75, and vanishes with long ($>150$) or short ($<20$) context windows. We surmise that for long ($>150$) context window, the preceding context information might already be adequate for detection, while for short ($<20$) context windows, the context, regardless whether it is unidirectional or bidirectional, might not contain enough information for detection. 
\paragraph{Model predictions on GPT-3 generated text}
We visualize the predictions of BERT-large (offline) model on GPT-3 generated text in Fig \ref{fig:model_viz}. According to the 2021 census instruments \footnote{\url{https://www.statcan.gc.ca/en/statistical-programs/instrument/3438_Q1_V6}}, some identified spans like ``greenhouse gas emission'' and ``complete enumeration'' are indeed not included in the census, we assume they are recognized due to the topic or knowledge irrelevance with the ``census of agriculture'' in the pretrained corpus. Interestingly, the detection model predicts the high hallucination risk on ``structures and buildings'', which has subtle differences with ``total greenhouse area including enclosed structures'' (included in the instruments). The case study demonstrates the potentials of our model in identifying hallucinated content in the actual outputs of large-scale pretrained models.

\section{Related Work}
\paragraph{Reference-based Hallucination Detection}
Apart from human verification \cite{chen2018fast}, researchers have developed effective reference-based methods which automatically detect hallucination in the generated text using statistical n-gram matching \cite{dhingra-etal-2019-handling,liu-etal-2019-towards-comprehensive}, edit distance heuristics \cite{zhou2020detecting}, natural language inference \cite{kryscinski2019evaluating,falke2019ranking}, information extraction \cite{zhang2019optimizing,goodrich2019assessing} or question answering \cite{scialom2019answers,eyal2019question,wang2020asking}. Our approach differs from them in that we investigate the reference-free hallucination detection scenario. 

To reduce hallucinations in the reference-based setting, researchers have applied iterative training \cite{nie-etal-2019-simple}, post editing \cite{dong-etal-2020-multi-fact}, soft constraints, e.g. attention manipulation \cite{kiddon-etal-2016-globally,sha2018order, hua-wang-2019-sentence,tian2019sticking,liu-etal-2019-towards-comprehensive} or optimal transport \cite{wang2020towards}, and template/scaffold guided schema with explicit plans \cite{ma-etal-2019-key,moryossef-etal-2019-step,balakrishnan-etal-2019-constrained,du-etal-2020-schema,liu2021towards}, e.g. text sequences which specify the narrative ordering, and implicit plans \cite{DBLP:conf/emnlp/WisemanSR18,ye2020variational,shen2020neural,li2020posterior}, e.g. (structured) hidden variables that corresponds to certain surface realization. 


\paragraph{Reference-free Detection Approaches}
Reference-free hallucination detection is closely related to 
fake news detection \cite{zellers2019neuralfakenews,zhou2020survey,DBLP:journals/corr/abs-2010-07475}, which aims to identify deliberate disinformation in a reference-free manner on social media and usually involves common-sense and world knowledge reasoning \cite{monti2019fake}, or fact checking \cite{thorne-etal-2018-fever}, where practitioners are asked to verify given claims without references by retrieving related evidence from Wikipedia.
Another line of research is to classify sentence-level language specificity \cite{li2015fast,gao2019predicting}, which scales from 1 (very general) - 5 (very specific) for short text, e.g. tweets, according to human annotation.

The proposed hallucination detection aims to examine the text in a finer granularity than fake news detection and fact checking. In the proposed task, most parts of the text remain faithful; our goal is to identify subtle hallucinations at the token-level. Fake news detection or specificity assessment, on the other hand, usually focus on sentence- or document-level detection. 



\section{Conclusions}
We have proposed a \emph{token-level reference-free} hallucination detection task and introduced a benchmark dataset \textsc{HaDes} for identifying fine granularity hallucination in free-form text generation. To create this dataset, we perturbed texts to simulate hallucination in NLG system, and performed an interative model-in-the-loop annotation approach to annotate the perturbed text in an imbalanced label scenario. We have further provided comprehensive analyses of \textsc{HaDes} and evaluated several baseline models to establish initial benchmarks. We hope that the proposed task and dataset will shed light on high-resolution hallucination detection in free-form text generation and will eventually lead to real-time hallucination prevention.

\section*{Broader Impact and Ethnic Consideration}
This study aims to facilitate the recognition of potential hallucinated content produced by large-scale pretrained models in the free-form generation. We support this goal with a novel \textit{reference-free, token-level} hallucination task and the corresponding annotated dataset \textsc{HaDeS}. The detection model trained with \textsc{HaDeS} could be potentially useful in both online and offline settings. 
For online settings it is possible to guide beam search or suppress the probability of hallucinated tokens through the detection models. For offline settings our system may expedite the human-in-the-loop post-examination in product deployment.

We design our model to detect hallucination to factual statement. The learned knowledge should be able to be transferred to other domain like social chatbot once the chat is regarding certain facts (e.g. a celebrity, a historical event). Wikipedia dataset covers a lot of facts, domains and topics, making it ideal for our study. We thus collect the \textsc{HaDeS} dataset from Wikipedia. All text on Wikipedia is licensed under the Creative Commons Attribution/Share-Alike 3.0 Unported License. During the annotation, all involved annotators voluntarily participated with decent payment. 

\section*{Acknowledgments}
The authors would like to thank the anonymous reviewers for their thoughtful and constructive comments.
Tianyu and Zhifang gratefully acknowledge the support of the National Key Research and Development Program of China 2020AAA0106701 and National Science Foundation of China project U19A2065.

\bibliography{acl2020}

\begin{thebibliography}{59}
\expandafter\ifx\csname natexlab\endcsname\relax\def\natexlab#1{#1}\fi

\bibitem[{Balakrishnan et~al.(2019)Balakrishnan, Rao, Upasani, White, and
  Subba}]{balakrishnan-etal-2019-constrained}
Anusha Balakrishnan, Jinfeng Rao, Kartikeya Upasani, Michael White, and Rajen
  Subba. 2019.
\newblock \href {https://doi.org/10.18653/v1/P19-1080} {Constrained decoding
  for neural {NLG} from compositional representations in task-oriented
  dialogue}.
\newblock In \emph{Proceedings of the 57th Annual Meeting of the Association
  for Computational Linguistics}, pages 831--844, Florence, Italy. Association
  for Computational Linguistics.

\bibitem[{Bird(2006)}]{bird2006nltk}
Steven Bird. 2006.
\newblock Nltk: the natural language toolkit.
\newblock In \emph{Proceedings of the COLING/ACL 2006 Interactive Presentation
  Sessions}, pages 69--72.

\bibitem[{Brown et~al.(2020)Brown, Mann, Ryder, Subbiah, Kaplan, Dhariwal,
  Neelakantan, Shyam, Sastry, Askell, Agarwal, Herbert-Voss, Krueger, Henighan,
  Child, Ramesh, Ziegler, Wu, Winter, Hesse, Chen, Sigler, Litwin, Gray, Chess,
  Clark, Berner, McCandlish, Radford, Sutskever, and
  Amodei}]{brown2020language}
Tom Brown, Benjamin Mann, Nick Ryder, Melanie Subbiah, Jared~D Kaplan, Prafulla
  Dhariwal, Arvind Neelakantan, Pranav Shyam, Girish Sastry, Amanda Askell,
  Sandhini Agarwal, Ariel Herbert-Voss, Gretchen Krueger, Tom Henighan, Rewon
  Child, Aditya Ramesh, Daniel Ziegler, Jeffrey Wu, Clemens Winter, Chris
  Hesse, Mark Chen, Eric Sigler, Mateusz Litwin, Scott Gray, Benjamin Chess,
  Jack Clark, Christopher Berner, Sam McCandlish, Alec Radford, Ilya Sutskever,
  and Dario Amodei. 2020.
\newblock \href
  {https://proceedings.neurips.cc/paper/2020/file/1457c0d6bfcb4967418bfb8ac142f64a-Paper.pdf}
  {Language models are few-shot learners}.
\newblock In \emph{Advances in Neural Information Processing Systems},
  volume~33, pages 1877--1901. Curran Associates, Inc.

\bibitem[{Center(2005)}]{center2005brier}
NOAA-CIRES Climate~Diagnostics Center. 2005.
\newblock Brier skill scores, rocs, and economic value diagrams can report
  false skill.

\bibitem[{Chen and Bansal(2018)}]{chen2018fast}
Yen-Chun Chen and Mohit Bansal. 2018.
\newblock \href {https://doi.org/10.18653/v1/P18-1063} {Fast abstractive
  summarization with reinforce-selected sentence rewriting}.
\newblock In \emph{Proceedings of the 56th Annual Meeting of the Association
  for Computational Linguistics (Volume 1: Long Papers)}, pages 675--686,
  Melbourne, Australia. Association for Computational Linguistics.

\bibitem[{Cohn et~al.(1996)Cohn, Ghahramani, and Jordan}]{cohn1996active}
David~A Cohn, Zoubin Ghahramani, and Michael~I Jordan. 1996.
\newblock Active learning with statistical models.
\newblock \emph{Journal of artificial intelligence research}, 4:129--145.

\bibitem[{Devlin et~al.(2019)Devlin, Chang, Lee, and
  Toutanova}]{devlin-etal-2019-bert}
Jacob Devlin, Ming-Wei Chang, Kenton Lee, and Kristina Toutanova. 2019.
\newblock \href {https://doi.org/10.18653/v1/N19-1423} {{BERT}: Pre-training of
  deep bidirectional transformers for language understanding}.
\newblock In \emph{Proceedings of the 2019 Conference of the North {A}merican
  Chapter of the Association for Computational Linguistics: Human Language
  Technologies, Volume 1 (Long and Short Papers)}, pages 4171--4186,
  Minneapolis, Minnesota. Association for Computational Linguistics.

\bibitem[{Dhingra et~al.(2019)Dhingra, Faruqui, Parikh, Chang, Das, and
  Cohen}]{dhingra-etal-2019-handling}
Bhuwan Dhingra, Manaal Faruqui, Ankur Parikh, Ming-Wei Chang, Dipanjan Das, and
  William Cohen. 2019.
\newblock \href {https://doi.org/10.18653/v1/P19-1483} {Handling divergent
  reference texts when evaluating table-to-text generation}.
\newblock In \emph{Proceedings of the 57th Annual Meeting of the Association
  for Computational Linguistics}, pages 4884--4895, Florence, Italy.
  Association for Computational Linguistics.

\bibitem[{Dong et~al.(2020)Dong, Wang, Gan, Cheng, Cheung, and
  Liu}]{dong-etal-2020-multi-fact}
Yue Dong, Shuohang Wang, Zhe Gan, Yu~Cheng, Jackie Chi~Kit Cheung, and Jingjing
  Liu. 2020.
\newblock \href {https://doi.org/10.18653/v1/2020.emnlp-main.749} {Multi-fact
  correction in abstractive text summarization}.
\newblock In \emph{Proceedings of the 2020 Conference on Empirical Methods in
  Natural Language Processing (EMNLP)}, pages 9320--9331, Online. Association
  for Computational Linguistics.

\bibitem[{Du et~al.(2020)Du, Oraby, Perera, Shen, Narayan-Chen, Chung,
  Venkatesh, and Hakkani-Tur}]{du-etal-2020-schema}
Yuheng Du, Shereen Oraby, Vittorio Perera, Minmin Shen, Anjali Narayan-Chen,
  Tagyoung Chung, Anushree Venkatesh, and Dilek Hakkani-Tur. 2020.
\newblock \href {https://www.aclweb.org/anthology/2020.inlg-1.35}
  {Schema-guided natural language generation}.
\newblock In \emph{Proceedings of the 13th International Conference on Natural
  Language Generation}, pages 283--295, Dublin, Ireland. Association for
  Computational Linguistics.

\bibitem[{Esp{\'\i}ndola and Ebecken(2005)}]{espindola2005extending}
Rog{\'e}rio~P Esp{\'\i}ndola and Nelson~FF Ebecken. 2005.
\newblock On extending f-measure and g-mean metrics to multi-class problems.
\newblock \emph{WIT Transactions on Information and Communication
  Technologies}, 35.

\bibitem[{Eyal et~al.(2019)Eyal, Baumel, and Elhadad}]{eyal2019question}
Matan Eyal, Tal Baumel, and Michael Elhadad. 2019.
\newblock \href {https://doi.org/10.18653/v1/N19-1395} {Question answering as
  an automatic evaluation metric for news article summarization}.
\newblock In \emph{Proceedings of the 2019 Conference of the North {A}merican
  Chapter of the Association for Computational Linguistics: Human Language
  Technologies, Volume 1 (Long and Short Papers)}, pages 3938--3948,
  Minneapolis, Minnesota. Association for Computational Linguistics.

\bibitem[{Falke et~al.(2019)Falke, Ribeiro, Utama, Dagan, and
  Gurevych}]{falke2019ranking}
Tobias Falke, Leonardo F.~R. Ribeiro, Prasetya~Ajie Utama, Ido Dagan, and Iryna
  Gurevych. 2019.
\newblock \href {https://doi.org/10.18653/v1/P19-1213} {Ranking generated
  summaries by correctness: An interesting but challenging application for
  natural language inference}.
\newblock In \emph{Proceedings of the 57th Annual Meeting of the Association
  for Computational Linguistics}, pages 2214--2220, Florence, Italy.
  Association for Computational Linguistics.

\bibitem[{Gao et~al.(2019)Gao, Zhong, Preo{\c{t}}iuc-Pietro, and
  Li}]{gao2019predicting}
Yifan Gao, Yang Zhong, Daniel Preo{\c{t}}iuc-Pietro, and Junyi~Jessy Li. 2019.
\newblock Predicting and analyzing language specificity in social media posts.
\newblock In \emph{Proceedings of the AAAI Conference on Artificial
  Intelligence}, volume~33, pages 6415--6422.

\bibitem[{Goodrich et~al.(2019)Goodrich, Rao, Liu, and
  Saleh}]{goodrich2019assessing}
Ben Goodrich, Vinay Rao, Peter~J Liu, and Mohammad Saleh. 2019.
\newblock Assessing the factual accuracy of generated text.
\newblock In \emph{Proceedings of the 25th ACM SIGKDD International Conference
  on Knowledge Discovery \& Data Mining}, pages 166--175.

\bibitem[{Guo et~al.(2020)Guo, Dai, Vrande{\v{c}}i{\'c}, and
  Al-Rfou}]{guo-etal-2020-wiki}
Mandy Guo, Zihang Dai, Denny Vrande{\v{c}}i{\'c}, and Rami Al-Rfou. 2020.
\newblock \href {https://www.aclweb.org/anthology/2020.lrec-1.297}
  {{W}iki-40{B}: Multilingual language model dataset}.
\newblock In \emph{Proceedings of the 12th Language Resources and Evaluation
  Conference}, pages 2440--2452, Marseille, France. European Language Resources
  Association.

\bibitem[{Holtzman et~al.(2020)Holtzman, Buys, Du, Forbes, and
  Choi}]{holtzman2019curious}
Ari Holtzman, Jan Buys, Li~Du, Maxwell Forbes, and Yejin Choi. 2020.
\newblock \href {https://openreview.net/forum?id=rygGQyrFvH} {The curious case
  of neural text degeneration}.
\newblock In \emph{International Conference on Learning Representations}.

\bibitem[{Hua and Wang(2019)}]{hua-wang-2019-sentence}
Xinyu Hua and Lu~Wang. 2019.
\newblock \href {https://doi.org/10.18653/v1/D19-1055} {Sentence-level content
  planning and style specification for neural text generation}.
\newblock In \emph{Proceedings of the 2019 Conference on Empirical Methods in
  Natural Language Processing and the 9th International Joint Conference on
  Natural Language Processing (EMNLP-IJCNLP)}, pages 591--602, Hong Kong,
  China. Association for Computational Linguistics.

\bibitem[{Jia and Liang(2017)}]{jia-liang-2017-adversarial}
Robin Jia and Percy Liang. 2017.
\newblock \href {https://doi.org/10.18653/v1/D17-1215} {Adversarial examples
  for evaluating reading comprehension systems}.
\newblock In \emph{Proceedings of the 2017 Conference on Empirical Methods in
  Natural Language Processing}, pages 2021--2031, Copenhagen, Denmark.
  Association for Computational Linguistics.

\bibitem[{Kiddon et~al.(2016)Kiddon, Zettlemoyer, and
  Choi}]{kiddon-etal-2016-globally}
Chlo{\'e} Kiddon, Luke Zettlemoyer, and Yejin Choi. 2016.
\newblock \href {https://doi.org/10.18653/v1/D16-1032} {Globally coherent text
  generation with neural checklist models}.
\newblock In \emph{Proceedings of the 2016 Conference on Empirical Methods in
  Natural Language Processing}, pages 329--339, Austin, Texas. Association for
  Computational Linguistics.

\bibitem[{Kingma and Ba(2015)}]{kingma2014adam}
Diederik~P. Kingma and Jimmy Ba. 2015.
\newblock \href {http://arxiv.org/abs/1412.6980} {Adam: A method for stochastic
  optimization}.
\newblock In \emph{ICLR (Poster)}.

\bibitem[{Kryscinski et~al.(2020)Kryscinski, McCann, Xiong, and
  Socher}]{kryscinski2019evaluating}
Wojciech Kryscinski, Bryan McCann, Caiming Xiong, and Richard Socher. 2020.
\newblock \href {https://doi.org/10.18653/v1/2020.emnlp-main.750} {Evaluating
  the factual consistency of abstractive text summarization}.
\newblock In \emph{Proceedings of the 2020 Conference on Empirical Methods in
  Natural Language Processing (EMNLP)}, pages 9332--9346, Online. Association
  for Computational Linguistics.

\bibitem[{Li and Nenkova(2015)}]{li2015fast}
Junyi Li and Ani Nenkova. 2015.
\newblock Fast and accurate prediction of sentence specificity.
\newblock In \emph{Proceedings of the AAAI Conference on Artificial
  Intelligence}, volume~29.

\bibitem[{Li and Rush(2020)}]{li2020posterior}
Xiang~Lisa Li and Alexander Rush. 2020.
\newblock \href {https://doi.org/10.18653/v1/2020.acl-main.243} {Posterior
  control of blackbox generation}.
\newblock In \emph{Proceedings of the 58th Annual Meeting of the Association
  for Computational Linguistics}, pages 2731--2743, Online. Association for
  Computational Linguistics.

\bibitem[{Liu et~al.(2019)Liu, Luo, Yang, Wu, Chang, and
  Sui}]{liu-etal-2019-towards-comprehensive}
Tianyu Liu, Fuli Luo, Pengcheng Yang, Wei Wu, Baobao Chang, and Zhifang Sui.
  2019.
\newblock \href {https://doi.org/10.18653/v1/P19-1600} {Towards comprehensive
  description generation from factual attribute-value tables}.
\newblock In \emph{Proceedings of the 57th Annual Meeting of the Association
  for Computational Linguistics}, pages 5985--5996, Florence, Italy.
  Association for Computational Linguistics.

\bibitem[{Liu et~al.(2021)Liu, Zheng, Chang, and Sui}]{liu2021towards}
Tianyu Liu, Xin Zheng, Baobao Chang, and Zhifang Sui. 2021.
\newblock Towards faithfulness in open domain table-to-text generation from an
  entity-centric view.
\newblock In \emph{Proceedings of the AAAI Conference on Artificial
  Intelligence}, volume~35, pages 13415--13423.

\bibitem[{Liu et~al.(2020)Liu, Ott, Goyal, Du, Joshi, Chen, Levy, Lewis,
  Zettlemoyer, and Stoyanov}]{liu2019roberta}
Yinhan Liu, Myle Ott, Naman Goyal, Jingfei Du, Mandar Joshi, Danqi Chen, Omer
  Levy, Mike Lewis, Luke Zettlemoyer, and Veselin Stoyanov. 2020.
\newblock \href {https://openreview.net/forum?id=SyxS0T4tvS} {Roberta: A
  robustly optimized bert pretraining approach}.

\bibitem[{Ma et~al.(2019)Ma, Yang, Liu, Li, Zhou, and Sun}]{ma-etal-2019-key}
Shuming Ma, Pengcheng Yang, Tianyu Liu, Peng Li, Jie Zhou, and Xu~Sun. 2019.
\newblock \href {https://doi.org/10.18653/v1/P19-1197} {Key fact as pivot: A
  two-stage model for low resource table-to-text generation}.
\newblock In \emph{Proceedings of the 57th Annual Meeting of the Association
  for Computational Linguistics}, pages 2047--2057, Florence, Italy.
  Association for Computational Linguistics.

\bibitem[{Maynez et~al.(2020)Maynez, Narayan, Bohnet, and
  McDonald}]{maynez-etal-2020-faithfulness}
Joshua Maynez, Shashi Narayan, Bernd Bohnet, and Ryan McDonald. 2020.
\newblock \href {https://doi.org/10.18653/v1/2020.acl-main.173} {On
  faithfulness and factuality in abstractive summarization}.
\newblock In \emph{Proceedings of the 58th Annual Meeting of the Association
  for Computational Linguistics}, pages 1906--1919, Online. Association for
  Computational Linguistics.

\bibitem[{Monti et~al.(2019)Monti, Frasca, Eynard, Mannion, and
  Bronstein}]{monti2019fake}
Federico Monti, Fabrizio Frasca, Davide Eynard, Damon Mannion, and Michael~M
  Bronstein. 2019.
\newblock Fake news detection on social media using geometric deep learning.
\newblock \emph{arXiv preprint arXiv:1902.06673}.

\bibitem[{Moryossef et~al.(2019)Moryossef, Goldberg, and
  Dagan}]{moryossef-etal-2019-step}
Amit Moryossef, Yoav Goldberg, and Ido Dagan. 2019.
\newblock \href {https://doi.org/10.18653/v1/N19-1236} {{S}tep-by-step:
  {S}eparating planning from realization in neural data-to-text generation}.
\newblock In \emph{Proceedings of the 2019 Conference of the North {A}merican
  Chapter of the Association for Computational Linguistics: Human Language
  Technologies, Volume 1 (Long and Short Papers)}, pages 2267--2277,
  Minneapolis, Minnesota. Association for Computational Linguistics.

\bibitem[{Nelson and Schreiber(1992)}]{nelson1992word}
Douglas~L Nelson and Thomas~A Schreiber. 1992.
\newblock Word concreteness and word structure as independent determinants of
  recall.
\newblock \emph{Journal of memory and language}, 31(2):237--260.

\bibitem[{Nie et~al.(2019)Nie, Yao, Wang, Pan, and Lin}]{nie-etal-2019-simple}
Feng Nie, Jin-Ge Yao, Jinpeng Wang, Rong Pan, and Chin-Yew Lin. 2019.
\newblock \href {https://doi.org/10.18653/v1/P19-1256} {A simple recipe towards
  reducing hallucination in neural surface realisation}.
\newblock In \emph{Proceedings of the 57th Annual Meeting of the Association
  for Computational Linguistics}, pages 2673--2679, Florence, Italy.
  Association for Computational Linguistics.

\bibitem[{Nie et~al.(2020)Nie, Williams, Dinan, Bansal, Weston, and
  Kiela}]{nie-etal-2020-adversarial}
Yixin Nie, Adina Williams, Emily Dinan, Mohit Bansal, Jason Weston, and Douwe
  Kiela. 2020.
\newblock \href {https://doi.org/10.18653/v1/2020.acl-main.441} {Adversarial
  {NLI}: A new benchmark for natural language understanding}.
\newblock In \emph{Proceedings of the 58th Annual Meeting of the Association
  for Computational Linguistics}, pages 4885--4901, Online. Association for
  Computational Linguistics.

\bibitem[{Pedregosa et~al.(2011)Pedregosa, Varoquaux, Gramfort, Michel,
  Thirion, Grisel, Blondel, Prettenhofer, Weiss, Dubourg
  et~al.}]{pedregosa2011scikit}
Fabian Pedregosa, Ga{\"e}l Varoquaux, Alexandre Gramfort, Vincent Michel,
  Bertrand Thirion, Olivier Grisel, Mathieu Blondel, Peter Prettenhofer, Ron
  Weiss, Vincent Dubourg, et~al. 2011.
\newblock Scikit-learn: Machine learning in python.
\newblock \emph{the Journal of machine Learning research}, 12:2825--2830.

\bibitem[{Press et~al.(2020)Press, Smith, and Lewis}]{press2020shortformer}
Ofir Press, Noah~A Smith, and Mike Lewis. 2020.
\newblock Shortformer: Better language modeling using shorter inputs.
\newblock \emph{arXiv preprint arXiv:2012.15832}.

\bibitem[{Radford et~al.(2019)Radford, Wu, Child, Luan, Amodei, and
  Sutskever}]{radford2019language}
Alec Radford, Jeffrey Wu, Rewon Child, David Luan, Dario Amodei, and Ilya
  Sutskever. 2019.
\newblock Language models are unsupervised multitask learners.
\newblock \emph{OpenAI blog}, 1(8):9.

\bibitem[{Rebuffel et~al.(2021)Rebuffel, Roberti, Soulier, Scoutheeten,
  Cancelliere, and Gallinari}]{rebuffel2021controlling}
Cl{\'e}ment Rebuffel, Marco Roberti, Laure Soulier, Geoffrey Scoutheeten,
  Rossella Cancelliere, and Patrick Gallinari. 2021.
\newblock Controlling hallucinations at word level in data-to-text generation.
\newblock \emph{arXiv preprint arXiv:2102.02810}.

\bibitem[{Rohrbach et~al.(2018)Rohrbach, Hendricks, Burns, Darrell, and
  Saenko}]{rohrbach-etal-2018-object}
Anna Rohrbach, Lisa~Anne Hendricks, Kaylee Burns, Trevor Darrell, and Kate
  Saenko. 2018.
\newblock \href {https://doi.org/10.18653/v1/D18-1437} {Object hallucination in
  image captioning}.
\newblock In \emph{Proceedings of the 2018 Conference on Empirical Methods in
  Natural Language Processing}, pages 4035--4045, Brussels, Belgium.
  Association for Computational Linguistics.

\bibitem[{Scialom et~al.(2019)Scialom, Lamprier, Piwowarski, and
  Staiano}]{scialom2019answers}
Thomas Scialom, Sylvain Lamprier, Benjamin Piwowarski, and Jacopo Staiano.
  2019.
\newblock \href {https://doi.org/10.18653/v1/D19-1320} {Answers unite!
  unsupervised metrics for reinforced summarization models}.
\newblock In \emph{Proceedings of the 2019 Conference on Empirical Methods in
  Natural Language Processing and the 9th International Joint Conference on
  Natural Language Processing (EMNLP-IJCNLP)}, pages 3246--3256, Hong Kong,
  China. Association for Computational Linguistics.

\bibitem[{See et~al.(2019)See, Pappu, Saxena, Yerukola, and
  Manning}]{see2019massively}
Abigail See, Aneesh Pappu, Rohun Saxena, Akhila Yerukola, and Christopher~D.
  Manning. 2019.
\newblock \href {https://doi.org/10.18653/v1/K19-1079} {Do massively pretrained
  language models make better storytellers?}
\newblock In \emph{Proceedings of the 23rd Conference on Computational Natural
  Language Learning (CoNLL)}, pages 843--861, Hong Kong, China. Association for
  Computational Linguistics.

\bibitem[{Sha et~al.(2018)Sha, Mou, Liu, Poupart, Li, Chang, and
  Sui}]{sha2018order}
Lei Sha, Lili Mou, Tianyu Liu, Pascal Poupart, Sujian Li, Baobao Chang, and
  Zhifang Sui. 2018.
\newblock Order-planning neural text generation from structured data.
\newblock In \emph{Thirty-Second AAAI Conference on Artificial Intelligence}.

\bibitem[{Shen et~al.(2020)Shen, Chang, Su, Niu, and Klakow}]{shen2020neural}
Xiaoyu Shen, Ernie Chang, Hui Su, Cheng Niu, and Dietrich Klakow. 2020.
\newblock \href {https://doi.org/10.18653/v1/2020.acl-main.641} {Neural
  data-to-text generation via jointly learning the segmentation and
  correspondence}.
\newblock In \emph{Proceedings of the 58th Annual Meeting of the Association
  for Computational Linguistics}, pages 7155--7165, Online. Association for
  Computational Linguistics.

\bibitem[{Thorne and Vlachos(2018)}]{thorne-vlachos-2018-automated}
James Thorne and Andreas Vlachos. 2018.
\newblock \href {https://www.aclweb.org/anthology/C18-1283} {Automated fact
  checking: Task formulations, methods and future directions}.
\newblock In \emph{Proceedings of the 27th International Conference on
  Computational Linguistics}, pages 3346--3359, Santa Fe, New Mexico, USA.
  Association for Computational Linguistics.

\bibitem[{Thorne et~al.(2018)Thorne, Vlachos, Christodoulopoulos, and
  Mittal}]{thorne-etal-2018-fever}
James Thorne, Andreas Vlachos, Christos Christodoulopoulos, and Arpit Mittal.
  2018.
\newblock \href {https://doi.org/10.18653/v1/N18-1074} {{FEVER}: a large-scale
  dataset for fact extraction and {VER}ification}.
\newblock In \emph{Proceedings of the 2018 Conference of the North {A}merican
  Chapter of the Association for Computational Linguistics: Human Language
  Technologies, Volume 1 (Long Papers)}, pages 809--819, New Orleans,
  Louisiana. Association for Computational Linguistics.

\bibitem[{Tian et~al.(2019)Tian, Narayan, Sellam, and
  Parikh}]{tian2019sticking}
Ran Tian, Shashi Narayan, Thibault Sellam, and Ankur~P Parikh. 2019.
\newblock Sticking to the facts: Confident decoding for faithful data-to-text
  generation.
\newblock \emph{arXiv preprint arXiv:1910.08684}.

\bibitem[{Wang et~al.(2020{\natexlab{a}})Wang, Cho, and Lewis}]{wang2020asking}
Alex Wang, Kyunghyun Cho, and Mike Lewis. 2020{\natexlab{a}}.
\newblock \href {https://doi.org/10.18653/v1/2020.acl-main.450} {Asking and
  answering questions to evaluate the factual consistency of summaries}.
\newblock In \emph{Proceedings of the 58th Annual Meeting of the Association
  for Computational Linguistics}, pages 5008--5020, Online. Association for
  Computational Linguistics.

\bibitem[{Wang and Sennrich(2020)}]{wang-sennrich-2020-exposure}
Chaojun Wang and Rico Sennrich. 2020.
\newblock \href {https://doi.org/10.18653/v1/2020.acl-main.326} {On exposure
  bias, hallucination and domain shift in neural machine translation}.
\newblock In \emph{Proceedings of the 58th Annual Meeting of the Association
  for Computational Linguistics}, pages 3544--3552, Online. Association for
  Computational Linguistics.

\bibitem[{Wang et~al.(2020{\natexlab{b}})Wang, Wang, An, Yu, and
  Chen}]{wang2020towards}
Zhenyi Wang, Xiaoyang Wang, Bang An, Dong Yu, and Changyou Chen.
  2020{\natexlab{b}}.
\newblock \href {https://doi.org/10.18653/v1/2020.acl-main.101} {Towards
  faithful neural table-to-text generation with content-matching constraints}.
\newblock In \emph{Proceedings of the 58th Annual Meeting of the Association
  for Computational Linguistics}, pages 1072--1086, Online. Association for
  Computational Linguistics.

\bibitem[{Wiseman et~al.(2018)Wiseman, Shieber, and
  Rush}]{DBLP:conf/emnlp/WisemanSR18}
Sam Wiseman, Stuart Shieber, and Alexander Rush. 2018.
\newblock \href {https://doi.org/10.18653/v1/D18-1356} {Learning neural
  templates for text generation}.
\newblock In \emph{Proceedings of the 2018 Conference on Empirical Methods in
  Natural Language Processing}, pages 3174--3187, Brussels, Belgium.
  Association for Computational Linguistics.

\bibitem[{Wolf et~al.(2020)Wolf, Debut, Sanh, Chaumond, Delangue, Moi, Cistac,
  Rault, Louf, Funtowicz, Davison, Shleifer, von Platen, Ma, Jernite, Plu, Xu,
  Le~Scao, Gugger, Drame, Lhoest, and Rush}]{wolf-etal-2020-transformers}
Thomas Wolf, Lysandre Debut, Victor Sanh, Julien Chaumond, Clement Delangue,
  Anthony Moi, Pierric Cistac, Tim Rault, Remi Louf, Morgan Funtowicz, Joe
  Davison, Sam Shleifer, Patrick von Platen, Clara Ma, Yacine Jernite, Julien
  Plu, Canwen Xu, Teven Le~Scao, Sylvain Gugger, Mariama Drame, Quentin Lhoest,
  and Alexander Rush. 2020.
\newblock \href {https://doi.org/10.18653/v1/2020.emnlp-demos.6} {Transformers:
  State-of-the-art natural language processing}.
\newblock In \emph{Proceedings of the 2020 Conference on Empirical Methods in
  Natural Language Processing: System Demonstrations}, pages 38--45, Online.
  Association for Computational Linguistics.

\bibitem[{Yang et~al.(2019)Yang, Dai, Yang, Carbonell, Salakhutdinov, and
  Le}]{yang2019xlnet}
Zhilin Yang, Zihang Dai, Yiming Yang, Jaime Carbonell, Russ~R Salakhutdinov,
  and Quoc~V Le. 2019.
\newblock \href
  {https://proceedings.neurips.cc/paper/2019/file/dc6a7e655d7e5840e66733e9ee67cc69-Paper.pdf}
  {Xlnet: Generalized autoregressive pretraining for language understanding}.
\newblock In \emph{Advances in Neural Information Processing Systems},
  volume~32. Curran Associates, Inc.

\bibitem[{Ye et~al.(2020)Ye, Shi, Zhou, Wei, and Li}]{ye2020variational}
Rong Ye, Wenxian Shi, Hao Zhou, Zhongyu Wei, and Lei Li. 2020.
\newblock \href {https://openreview.net/forum?id=HkejNgBtPB} {Variational
  template machine for data-to-text generation}.
\newblock In \emph{International Conference on Learning Representations}.

\bibitem[{Zellers et~al.(2018)Zellers, Bisk, Schwartz, and
  Choi}]{zellers-etal-2018-swag}
Rowan Zellers, Yonatan Bisk, Roy Schwartz, and Yejin Choi. 2018.
\newblock \href {https://doi.org/10.18653/v1/D18-1009} {{SWAG}: A large-scale
  adversarial dataset for grounded commonsense inference}.
\newblock In \emph{Proceedings of the 2018 Conference on Empirical Methods in
  Natural Language Processing}, pages 93--104, Brussels, Belgium. Association
  for Computational Linguistics.

\bibitem[{Zellers et~al.(2019)Zellers, Holtzman, Rashkin, Bisk, Farhadi,
  Roesner, and Choi}]{zellers2019neuralfakenews}
Rowan Zellers, Ari Holtzman, Hannah Rashkin, Yonatan Bisk, Ali Farhadi,
  Franziska Roesner, and Yejin Choi. 2019.
\newblock \href
  {http://papers.nips.cc/paper/9106-defending-against-neural-fake-news.pdf}
  {Defending against neural fake news}.
\newblock In H.~Wallach, H.~Larochelle, A.~Beygelzimer, F.~d\textquotesingle
  Alch\'{e}-Buc, E.~Fox, and R.~Garnett, editors, \emph{Advances in Neural
  Information Processing Systems 32}, pages 9054--9065. Curran Associates, Inc.

\bibitem[{Zhang et~al.(2020)Zhang, Merck, Tsai, Manning, and
  Langlotz}]{zhang2019optimizing}
Yuhao Zhang, Derek Merck, Emily Tsai, Christopher~D. Manning, and Curtis
  Langlotz. 2020.
\newblock \href {https://doi.org/10.18653/v1/2020.acl-main.458} {Optimizing the
  factual correctness of a summary: A study of summarizing radiology reports}.
\newblock In \emph{Proceedings of the 58th Annual Meeting of the Association
  for Computational Linguistics}, pages 5108--5120, Online. Association for
  Computational Linguistics.

\bibitem[{Zhong et~al.(2020)Zhong, Tang, Xu, Wang, Duan, Zhou, Wang, and
  Yin}]{DBLP:journals/corr/abs-2010-07475}
Wanjun Zhong, Duyu Tang, Zenan Xu, Ruize Wang, Nan Duan, Ming Zhou, Jiahai
  Wang, and Jian Yin. 2020.
\newblock \href {https://doi.org/10.18653/v1/2020.emnlp-main.193} {Neural
  deepfake detection with factual structure of text}.
\newblock In \emph{Proceedings of the 2020 Conference on Empirical Methods in
  Natural Language Processing (EMNLP)}, pages 2461--2470, Online. Association
  for Computational Linguistics.

\bibitem[{Zhou et~al.(2021)Zhou, Neubig, Gu, Diab, Guzm{\'a}n, Zettlemoyer, and
  Ghazvininejad}]{zhou2020detecting}
Chunting Zhou, Graham Neubig, Jiatao Gu, Mona Diab, Francisco Guzm{\'a}n, Luke
  Zettlemoyer, and Marjan Ghazvininejad. 2021.
\newblock \href {https://doi.org/10.18653/v1/2021.findings-acl.120} {Detecting
  hallucinated content in conditional neural sequence generation}.
\newblock In \emph{Findings of the Association for Computational Linguistics:
  ACL-IJCNLP 2021}, pages 1393--1404, Online. Association for Computational
  Linguistics.

\bibitem[{Zhou and Zafarani(2020)}]{zhou2020survey}
Xinyi Zhou and Reza Zafarani. 2020.
\newblock A survey of fake news: Fundamental theories, detection methods, and
  opportunities.
\newblock \emph{ACM Computing Surveys (CSUR)}, 53(5):1--40.

\end{thebibliography}
\bibliographystyle{acl_natbib}
\clearpage
\newpage

\appendix

\section{Detailed Statistical Analysis}
In Table \ref{tab:stats_feature_detailed}, we provide detailed statistical analyses for different POS and NER tags in the \textsc{HaDes} dataset.
 Although the average word probability and average word entropy features differ among POS/NER tags, 
hallucinated content typically associates with higher word probability and word entropy irrespective of POS/NER tag. Strong correlation between hallucination labels and TF-IDF or PPMI features is not observed. 
\begin{table*}[]
    \centering
    \small
    \begin{tabular}{l|cc|cc|cc|cc}
    \hline
    \multirow{2}{*}{Tag} & \multicolumn{2}{c|}{Word Prob($\times1e^{-8}$)} &
    \multicolumn{2}{c|}{Entropy} & \multicolumn{2}{c|}{TF-IDF} &
    \multicolumn{2}{c}{PPMI} \\
    & $\mathcal{H}$ & $\mathcal{N}$ & $\mathcal{H}$ & $\mathcal{N}$
    & $\mathcal{H}$ & $\mathcal{N}$ & $\mathcal{H}$ & $\mathcal{N}$\\\hline
    POS:NOUN & $\mathbf{6.98}_{32.0}$ & $\mathbf{1.68}_{6.34}$ & $\mathbf{2.75}_{1.52}$ & $\mathbf{1.86}_{1.13}$ &
    $\mathbf{.025}_{.021}$ & $\mathbf{.023}_{.018}$ &
    $\mathbf{.213}_{.145}$ & $\mathbf{.228}_{.140}$ \\
    POS:VERB & $\mathbf{2.51}_{9.33}$ & $\mathbf{0.69}_{2.89}$ &
    $\mathbf{2.25}_{1.25}$ & $\mathbf{1.76}_{1.00}$ &
    $\mathbf{.019}_{.012}$ & $\mathbf{.018}_{.011}$ &
    $\mathbf{.206}_{.112}$ & $\mathbf{.216}_{.119}$ \\
    POS:ADJ & $\mathbf{8.16}_{44.8}$ & $\mathbf{2.86}_{18.9}$ &
    $\mathbf{2.95}_{1.46}$ & $\mathbf{2.38}_{1.23}$ &
    $\mathbf{.021}_{.017}$ & $\mathbf{.017}_{.009}$ &
    $\mathbf{.180}_{.128}$ & $\mathbf{.164}_{.117}$ \\
    POS:ADV & $\mathbf{5.13}_{14.2}$ & $\mathbf{2.65}_{12.2}$ &
    $\mathbf{2.56}_{1.18}$ & $\mathbf{1.97}_{1.09}$ & 
    $\mathbf{.016}_{.011}$ & $\mathbf{.014}_{.008}$ & 
    $\mathbf{.181}_{.114}$ & $\mathbf{.182}_{.105}$ \\
    POS:PROPN & $\mathbf{14.3}_{33.6}$ & $\mathbf{4.35}_{17.8}$ &
    $\mathbf{3.12}_{1.73}$ & $\mathbf{1.56}_{1.39}$ &
    $\mathbf{.029}_{.026}$ & $\mathbf{.033}_{.029}$ &
    $\mathbf{.198}_{.150}$ & $\mathbf{.312}_{.275}$\\
    POS:other & $\mathbf{9.56}_{31.1}$ & $\mathbf{3.28}_{15.7}$ &
    $\mathbf{2.64}_{1.61}$ & $\mathbf{1.26}_{0.97}$ &
    $\mathbf{.013}_{.013}$ & $\mathbf{.011}_{.010}$ &
    $\mathbf{.158}_{.107}$ & $\mathbf{.205}_{.092}$ \\ 
    NER:null & $\mathbf{5.37}_{25.6}$ & $\mathbf{1.24}_{7.19}$ &
    $\mathbf{2.52}_{1.47}$ & $\mathbf{1.79}_{1.06}$ &
    $\mathbf{.021}_{.019}$ & $\mathbf{.019}_{.014}$ &
    $\mathbf{.200}_{.132}$ & $\mathbf{.215}_{.126}$ \\
    NER:other & $\mathbf{8.43}_{25.4}$ & $\mathbf{5.06}_{21.5}$ &
    $\mathbf{2.93}_{1.56}$ & $\mathbf{1.65}_{1.44}$ &
    $\mathbf{.023}_{.023}$ & $\mathbf{.026}_{0.024}$ &
    $\mathbf{.189}_{.146}$ & $\mathbf{.263}_{.237}$ \\
    All & $\mathbf{5.85}_{25.6}$ & $\mathbf{1.30}_{7.67}$ &
    $\mathbf{2.58}_{1.49}$ & $\mathbf{1.78}_{1.07}$ &
    $\mathbf{.021}_{.019}$ & $\mathbf{.019}_{0.014}$ &
    $\mathbf{.198}_{.144}$ & $\mathbf{.216}_{.129}$ \\\hline
    \end{tabular}
    \caption{Detailed statistical features ($\mathbf{Mean}_{\mathbf{std}}$) for ``hallucinated'' ($\mathcal{H}$) and ``not hallucinated'' ($\mathcal{N}$) cases.}
    \label{tab:stats_feature_detailed}
\end{table*}

\begin{figure*}
\begin{center}
\includegraphics[width=0.95\linewidth]{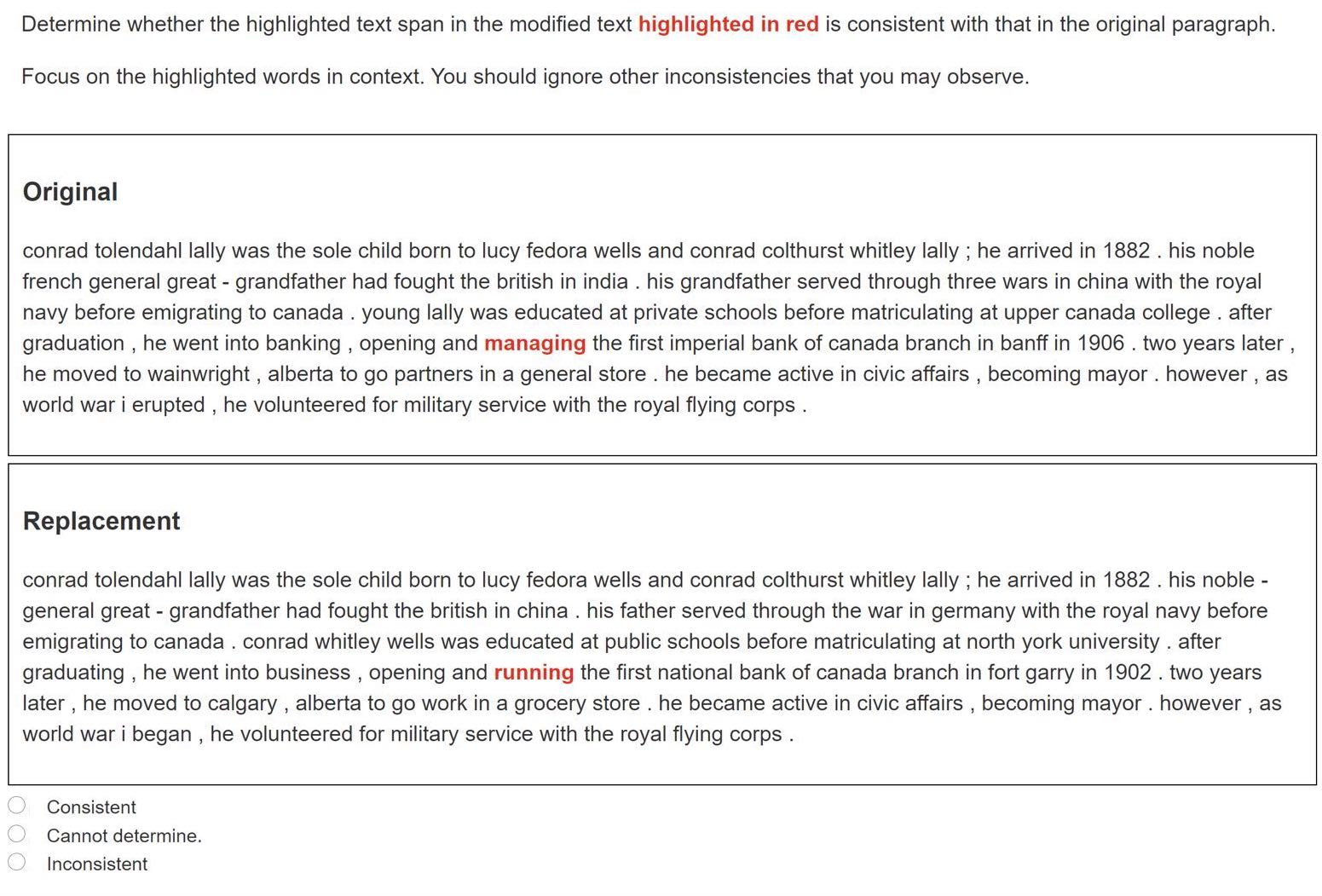}
\end{center}
\caption{The annotation interface for the proposed hallucination detection task.}
\label{fig:interface}
\end{figure*}

\section{Annotation Interface}
The annotation interface is provided in Fig \ref{fig:interface}. 

Note that throughout the annotation process we choose to involve an even number of, e.g. 4 or 6, annotators (Sec \ref{sec:data_annotation}) for an instance. The reason is that, we manage to involve extra annotators for controversial cases. If we pick an odd number of, e.g. 5 rather than 4, annotators, for each datapoint (binary classification) all possible results would be 0:5/1:4/2:3/3:2/4:1/5:0 in terms of the ratio of Hallucination/Consistent labels, which means no more annotators would be involved as they always reach consensus (majority wins).

\section{Subsampling Ratio For Label Rebalance}
We adopt an iterative model-in-the-loop method in data annotation. Since observe a label imbalance between  ``hallucination'' ($\mathcal{H}$) and ``not hallucination'' ($\mathcal{N}$) in the initial rounds of annotation,
we employ subsampling to rebalance the label distribution in Sec \ref{sec:data_annotation}.
We accumulate the data annotated in the all previous rounds, and train a detection model using the accumulated data. Then we apply the detection model to the unannotated data in the candidate data pool in order to select next batch of data as elaborated in Sec \ref{sec:data_annotation}.

We assume that the human annotation for $\mathcal{H}$ and $\mathcal{N}$ cases is the oracle, indicating actual $\mathcal{H}$/$\mathcal{N}$. Since the actual ``hallucinated'' is dominant, we try to subsample from the instances that are predicted by the detection model to be $\mathcal{H}$, in order to even out the distribution of actual $\mathcal{H}$/$\mathcal{N}$. To do this, we estimate the true positive rate\footnote{Defining $\mathcal{H}$ as the positive class.} (TPR, $\alpha$), true negative rate (TNR, $\beta$) and true precision ($\gamma$) of the detection model based on the annotations from the previous rounds. 

\begin{align}
    \text{TPR}=\frac {\text{TP}} {( \text{TP} + \text{FN} )} \triangleq \alpha
\end{align}

\begin{align}
    \text{TNR}=\frac {\text{TN}} {( \text{TN} + \text{FP} )} \triangleq \beta
\end{align}

\begin{align}
    \text{precision}=\frac {\text{TP}} {( \text{TP} + \text{FP} )} \triangleq \gamma
\label{eq:precision}
\end{align}

Where TP, FP, TN, FN are the abbreviations of ``true positive'', ``false positive'', ``true negative'' and ``false negative'' cases. 
We aim to subsample from the instances that are predicted as $\mathcal{H}$ from the detection model (TP + FP) with a subsampling ratio $s$, so that the actual $\mathcal{H}$ (TP + FN) is roughly equal to actual $\mathcal{N}$ (FP + TN) after the resampling. 
We denote TP and TN as $x$ and $y$ and represent FN and FP with $x,y,\alpha,\gamma,\beta$:

\begin{align}
    \text{FN}=\frac {1-\alpha} {\alpha} x
\end{align}
\begin{align}
    \text{FP}=\frac {1-\beta} {\beta} y
\end{align}

By substituting FN, FP into Eq. \eqref{eq:precision}, we have:

\begin{align}
    \gamma = \frac {x} {x + \frac{1-\beta}{\beta} y} 
\label{eq:gamma}
\end{align}

To make the distribution of actual $\mathcal{H}$/$\mathcal{N}$ even ($s$TP+FN=$s$FP+TN), we have:

\begin{align}
    sx + \frac {1-\alpha} {\alpha} x = s\frac {1-\beta} {\beta} y + y
\label{eq:s}
\end{align}

By combining Eq. \eqref{eq:gamma} and Eq. \eqref{eq:s}, we figure out the optimal subsampling ratio $s^*$.

\begin{align}
    s^* = \frac{-2\alpha\beta\gamma + \alpha\beta + \beta\gamma + \alpha\gamma-\gamma}{(2\gamma-1)\alpha(1-\beta)}
\end{align}

\end{document}